\documentclass[letterpaper, 10 pt, conference]{ieeeconf}  % Comment this line out if you need a4paper

\IEEEoverridecommandlockouts                              % This command is only needed if 
\usepackage{graphics} % for pdf, bitmapped graphics files
\usepackage{mathptmx} % assumes new font selection scheme installed
\usepackage{times} % assumes new font selection scheme installed
\usepackage{amsmath} % assumes amsmath package installed
\usepackage{amssymb}  % assumes amsmath package installed
\usepackage{subfig}
\usepackage[usenames, dvipsnames]{color}
\usepackage{soul}

\usepackage[final]{changes}
\usepackage{url}

\usepackage{graphicx}
\usepackage{caption}

\usepackage{algorithm}
\usepackage{algpseudocode}

\title{\LARGE \bf Low-level Active Visual Navigation: Increasing robustness of vision-based localization using potential fields}
\author{R\^{o}mulo T. Rodrigues,  Meysam Basiri, A. Pedro Aguiar, and Pedro Miraldo % <-this % stops a space
\thanks{R. T. Rodrigues,  M. Basiri, P. Miraldo are with the {\it Institute for Systems and Robotics} of the {\it Instituto Superior T\'{e}cnico}, {\it Universidade de Lisboa, Portugal}. A. P. Aguiar is with {\it  Faculty of Engineering},  {\it University of Porto, Portugal}. \newline
    E-Mail: {\tt\small romulo.rodrigues@tecnico.ulisboa.pt}}
}

\makeatletter
\renewcommand{\maketag@@@}[1]{\hbox{\m@th\normalsize\normalfont#1}}%
\makeatother           

\begin{document}

\maketitle
\thispagestyle{empty}
\pagestyle{empty}

%%%%%%%%%%%%%%%%%%%%%%%%%%%%%%%%%%%%%%%%%%%%%%%%%%%%%%%%%%%%%%%%%%%%%%%%%%%%%%%%
\begin{abstract}
This paper proposes a low level visual navigation algorithm to improve visual localization of a mobile robot. The algorithm, based on artificial potential fields, associates each feature in the current image frame with an attractive or neutral potential energy, with the objective of generating a control action that drives the vehicle towards the goal, while still favoring feature rich areas within a local scope, \replaced{thus improving}{improving in this way} the localization performance. One key property of the proposed method is that it does not rely on mapping, and therefore it is a lightweight solution that can be deployed on miniaturized aerial robots, in which memory and computational power are major constraints. Simulations and real experimental results using a mini quadrotor equipped with a downward looking camera demonstrate that the proposed method can effectively drive the vehicle to \replaced{a designated}{the} goal through a path that prevents localization failure.

\end{abstract}

%%%%%%%%%%%%%%%%%%%%%%%%%%%%%%%%%%%%%%%%%%%%%%%%%%%%%%%%%%%%%%%%%%%%%%%%%%%%%%%%
\section{INTRODUCTION}
% ################################################### %
% -------------- What is localization? -------------- %
% -------------- Is it important? Why? -------------- %
% ################################################### %
The problem of estimating the pose of a robot with respect to its environment, usually denoted as the localization problem, is one of the key challenges imposed on the deployment of autonomous Micro Aerial Vehicles (MAVs) in GPS-denied environments. Also addressed as state estimation, localization is of utmost importance, as the performance of other primary tasks such as autonomous navigation and obstacle avoidance relies on its accuracy. Beside compromising the mission, localization failure is potentially dangerous for humans and buildings in the vicinity, and the robot itself.

\begin{figure}[t!]
  \centering
  \subfloat[]{\includegraphics[width=0.98\linewidth]{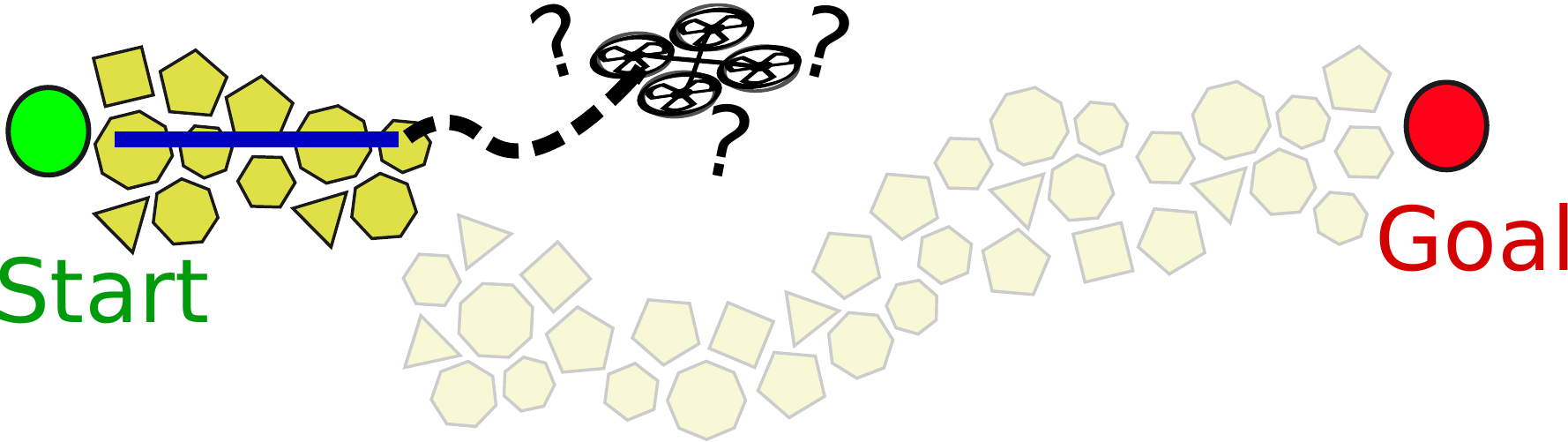} \label{fig:slam_fails}}\\
  \subfloat[]{\includegraphics[width=0.98\linewidth]{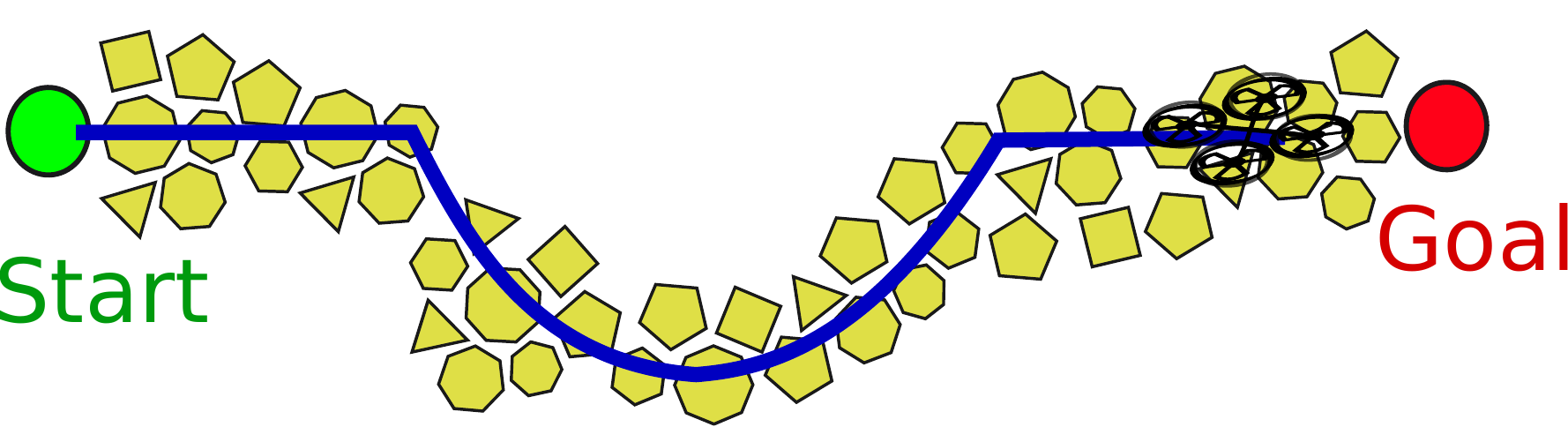}
  \label{fig:slam_works}}
    \caption{Fig.~\protect\subref{fig:slam_fails} \replaced{illustrates the}{shows} problem of \deleted{the} navigation without taking into consideration \added{features in} the environment, which can lead to localization failure. Fig.~\protect\subref{fig:slam_works} \replaced{presents the}{The} proposed solution \added{that} drives the vehicle towards a feature-rich area, enhancing localization robustness.}
    \label{uncertainty}
\end{figure}

% ################################################### %
% ----Is the localization problem solved? ------ %%
% ################################################### %
MAV localization in GPS-denied environments can be achieved with the aid of external systems such as motion tracking cameras. However, since such systems are not always available or practical, much research effort has been devoted \replaced{to the development of alternative systems relying on the use of}{in using} lightweight onboard sensors. The use of vision-based approaches for MAV localization has been the subject of many research works lately, e.g. \cite{loianno16, c1, c7}. \deleted{Most}Effective solutions combine visual estimation algorithms and inertial data within a filtering framework \cite{klein07, c3,mur-artal15,c5}. In particular, recent solutions such as ORB-SLAM \cite{mur-artal15} and SVO \cite{c3} are accurate in cluttered, feature rich environments and high texture scenarios. However, when visual cues are not available, the vision pipeline fails and the algorithm must be reinitialized. Moreover, during the re-initialization step state estimation relies on model propagation and inertial data, compromising localization and map consistency. If visual cues are not detected within few seconds, state estimation error and uncertainty grow fast and the vehicle may get lost, as is illustrated in Fig.~\ref{uncertainty}\subref{fig:slam_fails}.
%
%When visual cues become available once again, pose estimation uncertainty growth drops down.  However, initialized 3D mapped points have a high uncertainty, which does not decrease until previous mapped features are re-detected, i.e. the loop closure has been detected.

% As shown in Fig.~\ref{fig:uncertainty}, the vehicle may get lost 
%as shown in Fig.~\ref{fig:uncertainty}\subref{fig:uncertainty_fast_growth}. When visual cues become available once again, pose estimation uncertainty growth drops down. However, uncertainty does not decrease until previous mapped features are re-detected, i.e. the loop closure has been detected. However, notice that in the latter case, memory is required.

% ################################################### %
% ----  Proposed solution  ---- %%
% ################################################### %

This paper proposes a low-level strategy for MAVs to improve \deleted{the locally} Visual Odometry (VO) performance and map consistency. %\st{Therefore, high level localization task, such as Simultaneous Localization And Mapping (SLAM) are also benefited.} 
The solution aims at guiding the vehicle through \replaced{paths}{a path} with high quality image features. The method builds \replaced{upon}{on top of} an Artificial Potential Field (APF) framework, in which each image feature is associated with a corresponding potential level. \replaced{This yields an additional  reference velocity to the vehicle's control loop aimed at biasing its motion towards a feature-rich region in the current image frame.}{The control loop considers an additional reference velocity that points towards a feature rich region in the current image frame.} Spatial goal and APF\added{-generated} velocities are combined such that the vehicle drives toward the goal, while avoiding texture-less and non-static features regions, as illustrated in Fig.~\ref{uncertainty}\subref{fig:slam_works}. This behaviour \replaced{has the potential to}{can} reduce the growth of state estimation uncertainty and improve the localization accuracy. In addition, since this method does not rely on a map, it will not experience the same type of errors as in active SLAM solutions and, most important, it can be integrated in a complementary fashion. The proposed solution is specially \replaced{appealing}{interesting} for small scale vehicles, e.g. \cite{wagter14}, with low memory and limited processing power constraints that \replaced{prevent}{restrain} large map storage, loop closure detection, and global pose optimization. For these systems, visual SLAM is constrained to a low number of keyframes. Once \added{the} allocated memory is full, previous keyframes are discarded as new frames arrive \cite{c1}. Yet, local map consistency is required for proper 3D world point initialization.

\replaced{The}{This} paper is organized as follows: Sec.~II \replaced{summarizes}{presents} related work regarding passive localization, active localization, and artificial potential fields. Sec.~III introduces \added{some} basic notation and definitions. \deleted{In} Sec.~IV \replaced{describes the new method proposed}{, the proposed method is presented}. Sec.~V presents simulation and experimental results. Finally, Sec.~VI \replaced{contains the}{addresses} final remarks. This paper builds upon and extends previous results that were presented in \cite{rodrigues17} by the authors. In particular, simulations validate the proposed algorithm in different scenarios, the effect of some parameters on the proposed is evaluated, the benefits of an auto-tuning technique are demonstrated, \added{and extensive experiments on a real world scenario demonstrate the significant improvement on VO and SLAM performance for specific cases using the proposed method.}

% Problem addressed in this paper, namely the fast grow of the uncertainty on the robot's localization
%[By implicitly include visual features in the control in the low level active visual localization, the problem of uncertainty in \protect\subref{fig:uncertainty_fast_growth} can be minimized.

%Considering poor or wrong data association in the visual odometry pipeline is inevitable due sensor modelling error, image noise, illumination variation, non-static elements and texture-less regions. 

%This problem can be addressed at a high level using local bundle adjustment and loop closure, i.e. revisiting mapped regions, which drastically reduce estimation error and uncertainty. However, restricted onboard processing power limits batch optimization rate and low memory hinder map storage for loop closure detection. 

% In contrast to passive localization, in which feature initialization is a result of navigation, the strategy addressed in this paper takes advantage of the active localization formulation

 %As suggested by \cite{c2}, considering poor or wrong data association is inevitable due sensor modelling error, image noise, illumination variation, non-static elements and texture-less regions. 

% \subsection{Notations}
% \subsection{Our Contributions}

\section{Literature Review}
\label{sec:literature_review}
This section provides a general overview of the foundations of the problem tackled in this work. Relevant solution regarding visual odometry, active localization, and artificial potential field are discussed.  

\subsection{Visual Odometry}
\replaced{VO}{Visual odometry} estimates the camera motion by taking into consideration the motion of tracked features in two consecutive frames. Thus, it is a dead-reckoning method, i.e. estimates relative, rather than absolute pose. In passive VO \cite{scaramuzza11}, interest image regions are initialized and tracked as a consequence of navigation. Recent solutions\added{, e.g.} \cite{c3,mur-artal15,c5}\added{,} have achieved impressive results. The keypoint method developed in \cite{mur-artal15} extracts features from salient image regions, to recover camera pose using epipolar geometry. The hybrid method proposed in \cite{c3} combines both features and photometric information at high frame rates for a robust pose estimation. The dense strategy in \cite{c5} exploits photometric error to recover camera pose, integrating lens and camera parameters. The two last methods operate directly over the pixel intensity, being therefore more robust and accurate in sparsely textured environment. However, as suggested in \cite{c5}, the performance depends on the camera model, frame rate, and environment brightness. Nonetheless, we are not aware of any VO that does not require a quasi-static scenario and texture for working properly. The method addressed in this paper assures these two underlying assumptions hold during flight.

\subsection{Active Localization}
Active localization adds visual information in the control loop, with the goal of minimizing pose uncertainty. The problem is addressed in the literature under different formulations, with significant overlapping: active-SLAM \cite{c8,vidal-calleja10,bryson08}, planning under uncertainty \cite{c13,c6}, or next-best-view \cite{sadat14,c11}. 

Davison and Murray \cite{c8} define the main goal in active feature selection as building a map of features, which helps localizing the robot rather than an end result in itself. Their solution is not active in the sense that it does not actuate on the camera path. Instead, it chooses the best features to fixate the camera's view. Vidal-Calleja et al. \cite{vidal-calleja10} extends the former solution proposing a control law that drives the camera to the location that maximizes the expected information gain. Based on information theory, features corresponding to the hardest prediction measurement are chosen. A formal approach that relates SLAM observability and vehicle motion is addressed by Bryson and Sukkarieh \cite{bryson08}. The proposed on-line path planner decides whether explore new regions or revisit known features to improve localization. Trajectories that excite locally unobservable modes are preferred. 

Considering a known map (given {\it a priori}), the navigation task can be improved by planning routes that favor texture-rich areas. Achtelik et al. \cite{c13} addressed a Rapid-exploring Random Belief Tree (RRBT) framework that incorporate MAV dynamics and pose uncertainty. The method fails in a non-static scenario, or under illumination changes.  In \cite{c6}, Rapid-exploring Random Trees (RRT*) is extended, to take into account the pose uncertainty. Most informative trajectories are selected using Fischer informative matrix. Previous non-mapped or non-static regions have impact on local edges and vertexes affected by new information. While local changes in the nodes handle small discrepancies between map and environment, these methods are not scalable since an offline path planner requires an accurate map. 

Sadat et al. \cite{sadat14} propose a scoring function that takes into account the expected number of features for a given camera's viewpoint, using a mesh of triangles. The cost function of the optimized path planer includes path length and expect number of features. Mostegel et al. \cite{c11} propose a set of measurements, including geometric point quality and recognition probability, to analyze the impact of possible camera motions, and avoid localization loss. The solution relies on SLAM key-frames data. A local planner assures the vehicle reaches the destination in unexplored maps.

\begin{figure}[t!]
  \centering
  \includegraphics[width=0.90\linewidth]{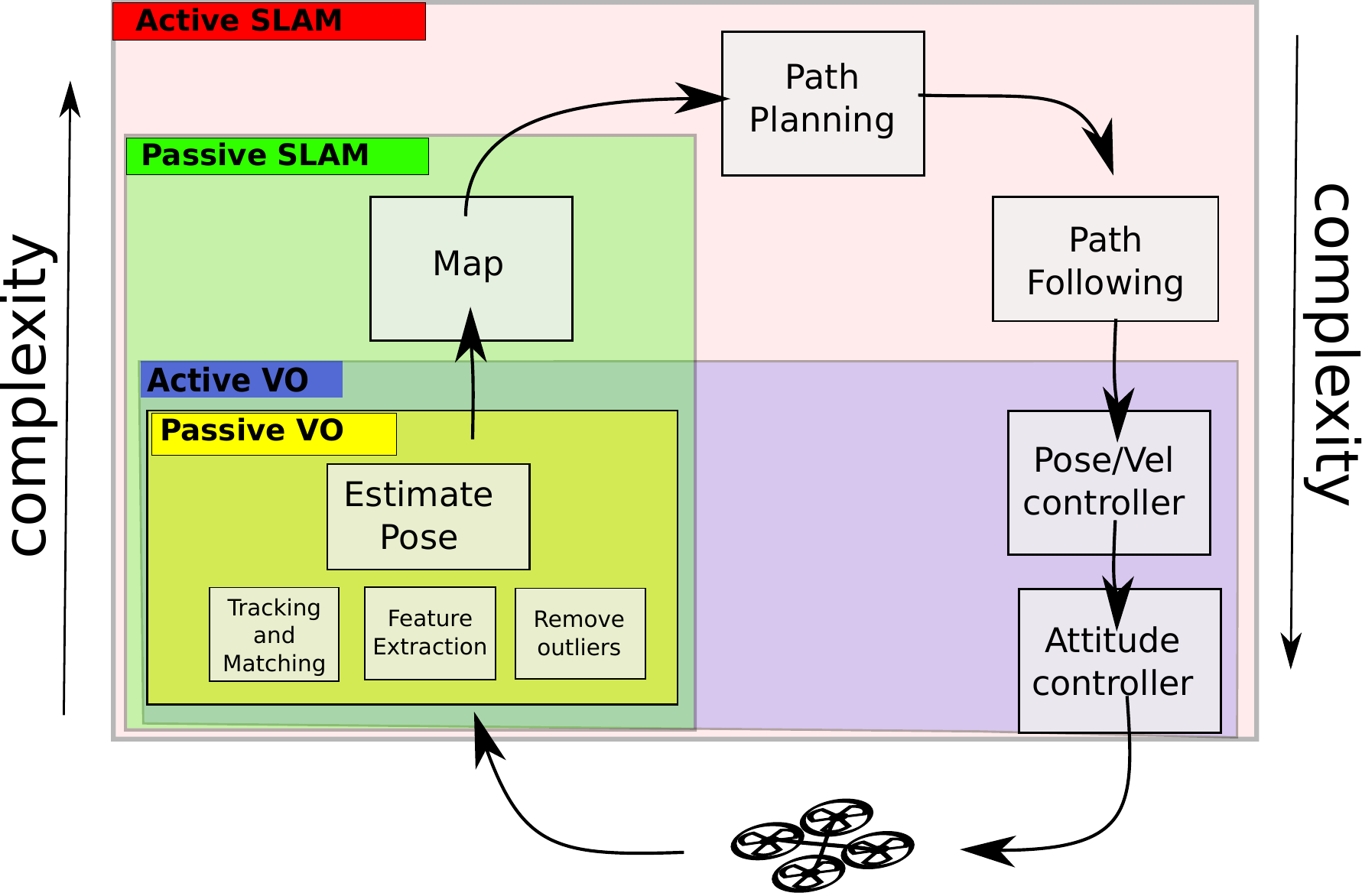}
  \caption{\replaced{Comparison between the method proposed in this paper (here denoted as Active VO) against most of the state-of-the-art alternatives (identified as Active SLAM).}{Passive and active localization  for mobile robots.}}
  \label{fig:active_vs_localization}
\end{figure}

\subsection{Artificial Potential fields}
Early work on artificial potential fields was done by Khatib \cite{khatib86} for real-time obstacle avoidance. In this set-up, the spatial goal was associated to an attractive force, while obstacles, detected on-the-fly, were associated to a repulsive force. In contrast to prior works that tackled obstacle avoidance as \added{a} high-level planning problem, APF proposed a low level real-time local solution. Furthermore, the author suggested integrating both high-level and low-level for better performance. The popularity of the potential field´s approach has grown steadily and, lately, it has found new applications besides obstacle avoidance, such as in navigation \cite{rimon92}, loop-closure detection \cite{vallve15} and mobile robot exploration \cite{jorge15}.
 
Fig.~\ref{fig:active_vs_localization}, inspired in \cite{siegwart04}, illustrates both active and passive localization frameworks. Most active localization solutions addressed in the literature are built on top of SLAM, and are therefore relatively complex and map dependent. In contrast, our method can be classified as active VO: it does not require a map, but only features selected as inliers in the current frame. 

\section{Notations and definitions}
\label{sec:notation_definition} 
Let $\{B\}$ and $\{I\}$ denote the 3D body fixed frame  attached to the vehicle and the 2D image plane frame attached to image plane, respectively. The origin of $\{B\}$ coincides with the center of gravity of the vehicle and the origin of $\{I\}$ corresponds to the top-left image pixel. Vectors are described in lower case bold and a leading superscript indicates its coordinate frame. The homogeneous coordinate of vector $\mathbf{v}$ is denoted as $\bar{\mathbf{v}}$. When a vector is described in $\{I\}$, the leading superscript is omitted. Matrices are written in upper case and sets in calligraphic letter. The transformation from body to image frame ${T}= K [{R}|\mathbf{t}] \in \mathbb{R}^{3\times 4}$ is known, where $R$ and $\mathbf{t}$ denote the rotation and translation from body frame to the downward looking camera, respectively, and K is the intrisic parameter matrix of the camera. For the sake of simplicity, the camera frame is omitted, as usually a static or a known transformation relates camera and body frames.

A high level positioning controller computes ${^B\mathbf{v}_{g}} = (v_x,v_y,0) \in \mathbb{R}^3$, a velocity \added{reference} that drives the vehicle to the spatial goal. Notice that only planar motion is considered, that is, the vehicle keeps a constant height. Also, let $\mathbf{p}=(u,v) \in  \mathbb{R}^2$ be the undistorted coordinates of an image feature \cite{hartley04}. In particular, $\mathcal{F}$ is the set of features tracked and selected as inliers in the current image frame.

\section{Low-level Active Visual Navigation}
\label{sec:proposed_method}
\begin{figure}[t!]
  \centering
  \includegraphics[width=0.9\linewidth]{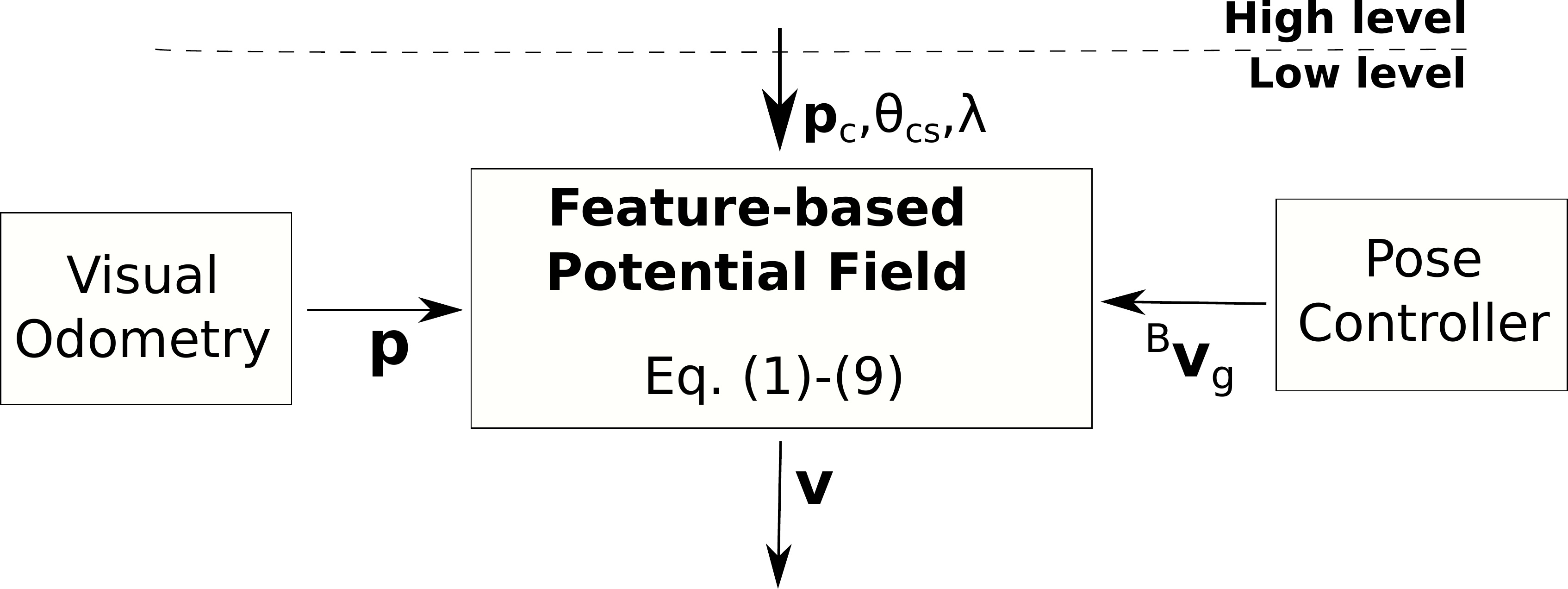}
  \caption{Block diagram representing the proposed APF framework, and its communication with the other modules.}
  \label{fig:solution_overview}
\end{figure}
This section describes the method proposed to compute the velocity reference that drives the vehicle towards a spatial goal, while avoiding low feature areas. This is achieved by adding a component to the goal velocity vector ${^B\mathbf{v}_{g}}$ so as to favor rich regions regarding features. A block diagram shown in Fig. \ref{fig:solution_overview} allows visualizing the information flow addressed in the remainder of this section. The proposed method is fed by a visual odometry algorithm, which provides inlier features, and a pose controller, e.g. path-following algorithm, that computes the goal velocity reference. The desired vehicle velocity is computed by the proposed low level active visual navigation method. A higher level system, such as a mission controller, may dynamically tune the parameters of the method proposed\added{, which will be presented and discussed next.}

\subsection{Features to charge}
To each feature, an attractive or neutral potential energy is associated. Associating similar potential energy to every feature in the image frame is not adequate, since the vehicle could be easily trapped at a local minimum or subject to sudden changes when new features are extracted. Instead, taking advantage of the fact that the camera provides bearing information, the proposed method considers the orientation of each feature w.r.t. the direction of $\mathbf{v}_g$, i.e. the projection of the goal velocity in the image frame. As expected in a potential field framework, the final goal itself plays a role in the local decision making process.
\begin{figure}[t!]
  \centering
  \includegraphics[width=0.6\linewidth]{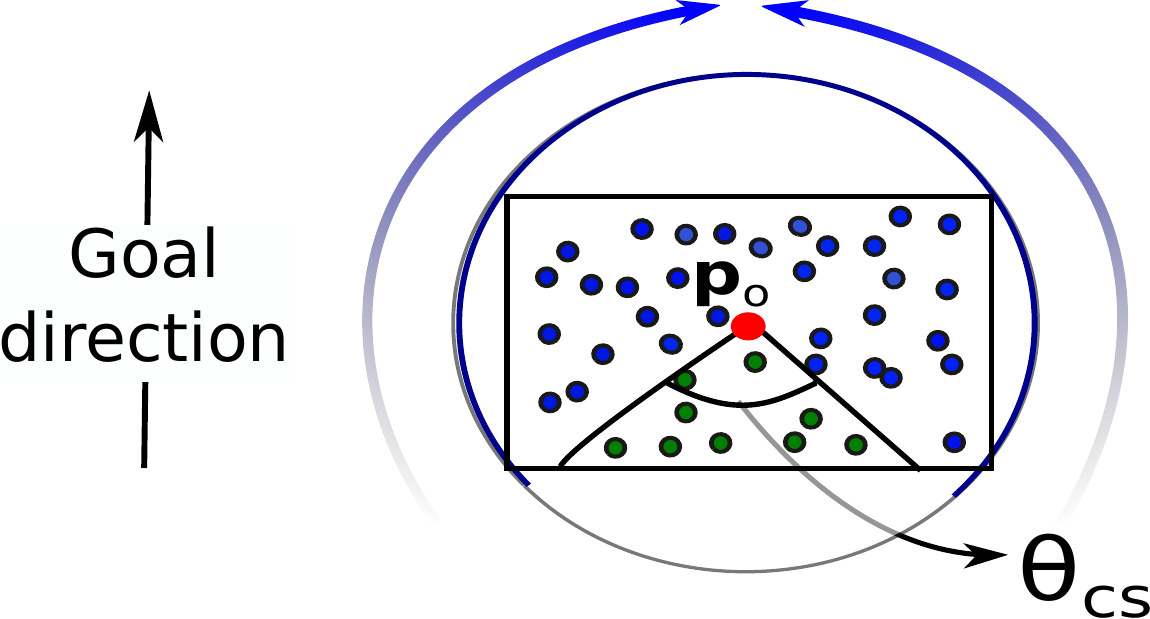}
  \caption{Attractive potential energy increase in the direction of the blue arrow. Charges are neutral within the circle segment defined by $\theta_{cs}$. Attractive and neutral charges are represented by blue and green dots, respectively.}
  \label{fig:energy_assignment}
\end{figure}
Let \replaced{$\mathbf{p}_o = (c_x, c_y)$ be the camera optical center obtained in the camera calibration step}{$\mathbf{p}_c = (u_r, v_r)$ be a point that belongs to the image frame}\deleted{, and consider that the feature based velocity shall be computed at that point}. For each feature ${\mathbf{p}_i} \in \mathcal{F}$, compute
\begin{align}
\hat{\mathbf{p}}_i &= \mathbf{p}_i -  \mathbf{p}_o \label{eq:p_hat}\\
\theta_i &= \arccos\left(\frac{\langle \hat{\mathbf{p}}_i, {\mathbf{v}_g} \rangle}{\|\hat{\mathbf{p}}_i\|\,\|{\mathbf{v}_g}\|}\right),
\label{eq:theta}
\end{align}
where $\theta_i \in [0,\pi]$ and ${\mathbf{v}}_g \sim {T}\, {^B\bar{\mathbf{v}}_{g}}$. Notice that only the direction of ${^B\mathbf{v}_g}$ is taken into account for computing the feature-based velocity vector\footnote{The last coordinate of its homogeneous form must be set to $0$.}.
\begin{figure*}[t!]
  \centering
  \includegraphics[width=0.85\textwidth,height=8cm,keepaspectratio]{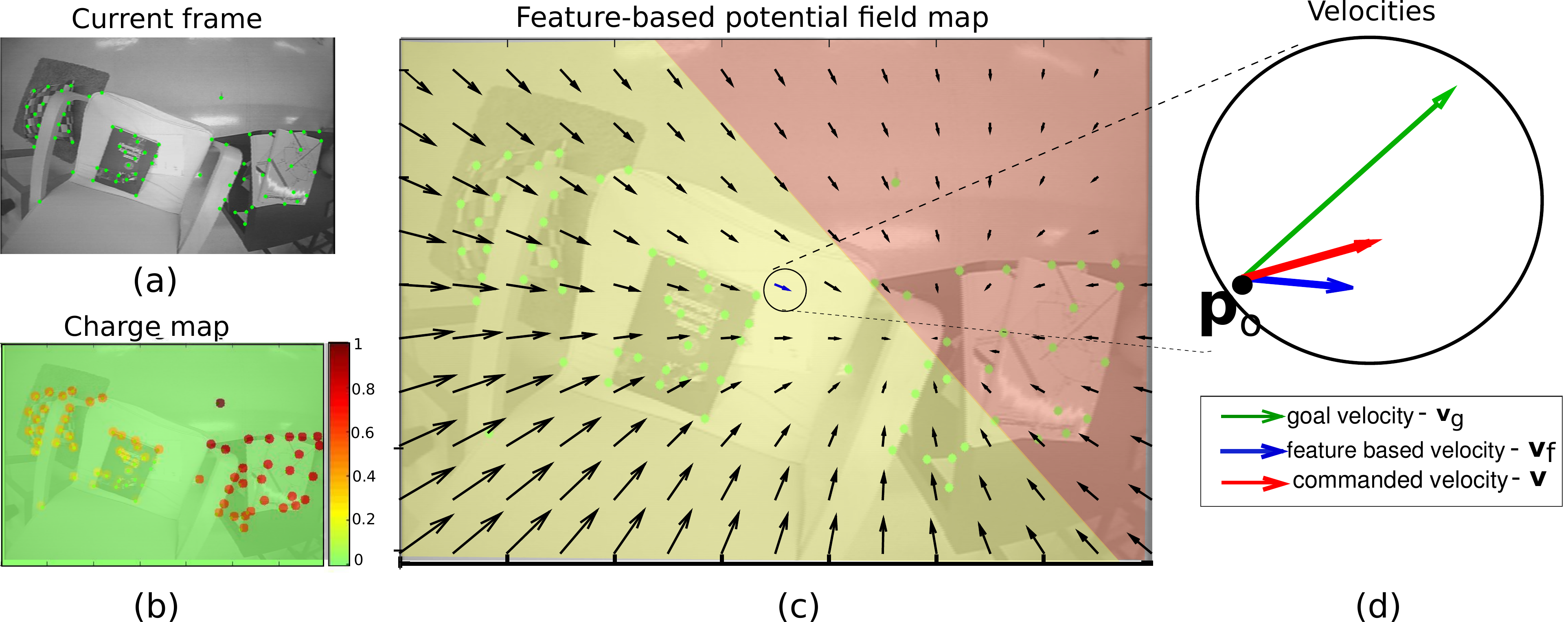}
  \caption{Case study for the feature based potential field. Current image frame and inlier features (a). Considering $\mathbf{p}_c$ as the optical center, the charge map is built and represented as a heat-map. (b). The potential field map shows the action derived when evaluating at different $\mathbf{p}_c$ (c). Goal-friendly region is shown in yellow and feature-friendly in red (more detail about these regions is shown in Sec.~\ref{sec:our_method_discussion}). (d) shows commanded reference velocity as the combination of goal and feature velocity. ($\mathbf{p}_c=\mathbf{p}_o$, $\lambda = 0.1, \theta_{CS} = 10^{o.}, d=50$ pixel, $s=150$ pixel).}
  \label{fig:showcase}
\end{figure*}
Without loss of generality, features in the image plane can be confined within the boundary of a circle centered at $\mathbf{p}_o$, as shown in Fig.~\ref{fig:energy_assignment}. Let the central angle $\hat{\theta}_{cs}$ define a circular segment in the circle and $\theta_{cs}=\pi - \hat{\theta}_{cs}/2$. A charge is represented as the tuple $Q_i=(\hat{\mathbf{p}}_i, q_i, {\theta_{cs}} )$, where $q_i \in [0,1]$ is its corresponding potential energy. The charging policy is defined below:
\begin{equation}
{q}_{_i} =
\arraycolsep=0.4pt\def\arraystretch{1.0}
\left \{ \begin{array}{ll}
1 - \frac{\theta_i}{\theta_{cs}} , & \text{ if } \theta_i \leq \theta_{cs} \\
0 , & \text{ if } \theta_i > \theta_{cs} \\
\end{array} \right. .
\label{eq:q_pos}
\end{equation}

Based on the current frame information, the considered charging policy associates high attractive potential energy to features localized in the goal direction. Features localized away from the goal direction, on the region defined by the circle segment, have a neutral charge.

\subsection{Vector Field}
Each charge $Q_i \in \mathcal{Q}$ exerts a force  $\mathbf{f}_{i}$ at a point $\mathbf{p}_c$ that belongs to the imag frame, given by
\begin{equation}
\mathbf{f}_{i} =
\arraycolsep=0.4pt\def\arraystretch{1.0}
\left \{ \begin{array}{ll}
(0,\ 0)^T , & \text{ if } d_i < r \\
\frac{(d_i-r)}{s}\ q_i\Big(\cos(\phi_i),\ \sin(\phi_i) \Big) ^T , & \text{ if } r \leq d_i \leq s + r \\
q_i\Big( \cos(\phi_i),\ \sin(\phi_i) \Big) ^T , & \text{ if } d > s + r
\end{array} \right. ,
\end{equation}
where $r$ is the distance in pixels that a charge must be from the evaluated point to exert any force on it; $s$ is the spread in pixels of the potential field, and $d_i$ and $\phi_i$ are computed as follow
\begin{align}
    d_i &=\|\mathbf{p}_i - \mathbf{p}_c\| \label{eq:d} \\
    \phi_i &= \sphericalangle(\mathbf{p}_i, \mathbf{p}_c).
\end{align}
The total force $\mathbf{f}$ on the point $\mathbf{p}_c$ can be computed as
%\begin{equation}
$\mathbf{f}~=~\displaystyle\sum_{i}^{}~\mathbf{f}_{i}$,
%\end{equation}
which can be normalized and transformed into a feature based velocity command ${\bar{\mathbf{v}}_{f}}$. Its direction in homogeneous coordinates is given as
\begin{equation}
{\bar{\mathbf{v}}_{f}} = \frac{1}{\|\mathbf{f}\|}
\begin{bmatrix}   
\mathbf{f} \\
0
\end{bmatrix},
\end{equation}
%ensuring that $\|\bar{\mathbf{v}}_{f}\| \leq 1$.

Finally, the proposed reference velocity takes the form
\begin{equation}
{\bar{\mathbf{v}}} = \lambda \bar{\mathbf{v}}_{g} + (1-\lambda)\bar{\mathbf{v}}_{f},
\label{eq:proposed_command} 
\end{equation}
where $\lambda$ is a weight factor and ${\bar{\mathbf{v}}_{g}}$ is a normalized velocity. The velocity ${\bar{\mathbf{v}}}$ can be transformed from the image frame to the body frame and scaled accordingly.

\subsection{Discussion}
\label{sec:our_method_discussion}

Artificial potential fields frameworks usually take into consideration the robot position when computing forces - obstacles exert repulsive force and the goal an attractive force. 
In the proposed framework, features can be either attractive or neutral accordingly to their position w.r.t. the point $\mathbf{p}_c$ being considered and the goal direction.

A case study is illustrated in Fig.~\ref{fig:showcase}. In particular, Fig.~\ref{fig:showcase}(a) shows a frame and extracted features classified as inliers. The goal velocity ${\mathbf{v}_g}$ is directed towards the top-right pixel in the image frame - a poor zone regarding the number of features. Figure ~\ref{fig:showcase}(b) depicts the potential energy associate with each feature when evaluating the action induced at the central pixel of the camera, $\mathbf{p}_o$. The potential field map (see Fig.~\ref{fig:showcase}(c)) shows the corresponding field for different values of $\mathbf{p}_c$. For each point in the map, \eqref{eq:p_hat}--\eqref{eq:proposed_command} must be computed. However, for visualization, $\mathbf{v}_{f}$ is not normalized. Observe that the map can be classified in two different zones, which are labelled as \textit{goal-friendly} and \textit{feature-friendly} actuation zones. Both regions are shown in the potential field map in the yellow (goal-friendly) and the red (feature-friendly) background. Suppose $\lambda = 0$; then, according to \eqref{eq:proposed_command} the vehicle follows $\mathbf{v}_{f}$. If $\mathbf{p}_c$ is within a feature-friendly region, the vehicle favors more the features than the goal.  As a matter of fact, the vehicle will move away from the goal. On the contrary, if $\mathbf{p}_c$ is within a goal-friendly region, the vehicle will move towards the goal. The absolute value of the angle between the goal velocity and the feature based velocity determines these regions -- an acute angle indicates a goal-friendly zone and an obtuse angle is related to a feature-friendly zone. The radius of the charge $r$ determines whether features close to the point $\mathbf{p}_c$ affect the solution or not. Meanwhile, the spread $s$ determines the strength of each charge. The larger the spread is, the more influence the charges away from the point evaluated will have on the feature driven action. Thus, it limits the prediction horizon based on the local frame, that is, the belief that a feature on an edge indicates that there will be more features on that direction.

\subsection{Auto-tuning}
\label{subsection:autotuning}
The value of $\lambda$ determines whether the vehicle shall follow the feature-based velocity or the goal velocity. The main motivation to follow the latter rather than the former is the possibility of localization failure. Therefore, the value of $\lambda$ shall be selected considering a given performance index of the localization algorithm being employed. In this paper, the number of inlier features tracked across consecutive frames ($N_f$) is considered. If the number of tracked features is above an upper threshold ($T_{f_{MAX}}$), $\lambda$ is set to $\lambda_{MAX}$. Conversely, if the number of tracked features is below a lower threshold ($T_{f_{MIN}}$), $\lambda$ is set to $\lambda_{MIN}$. If the number of tracked features is in between these two thresholds, then $\lambda$ is proportional to the number of features. The auto-tuning method is summarized as
\begin{equation}
\lambda(N_f) =
\arraycolsep=0.4pt\def\arraystretch{1.0}
\left \{ \begin{array}{ll}
\lambda_{MIN} , & \text{ if } N_f < T_{f_{MIN}} \\
\alpha N_f + \beta, & \text{ if } T_{f_{MIN}} \leq N_f \leq T_{f_{MAX}} \\
\lambda_{MAX} , & \text{ if } N_f > T_{f_{MAX}}
\end{array} \right. ,
\end{equation}
where $\alpha$ and $\beta$ are constants such that $\lambda(N_f)$ is continuous.

\section{Results}

This section presents simulation results and real world experiments for multiple flight tests. For both simulation and real world experiments, images and commands are published and received within the ROS (Robot Operating System) environment\added{~\cite{ros}}. The feature-based potential field runs online as the decision making process, sending desired velocity commands to the velocity controller. A first order low pass filter, with a cutoff frequency at $20$~Hz, provides smooth control reference for the feature driven action. When a new image frame is received, Shi-Tomasi features \cite{shi94} are extracted. In addition to a minimum quality threshold, only $100$ features with the highest response are selected. Then, features are tracked across two consecutive frames using the Lucas-Kanade Tracker (LKT) \cite{lucas81}. By resorting to a RANSAC \cite{fischler81} framework, the 8-point algorithm classifies features as inliers or outliers. The solution is robust in the presence of few false inliers, such that a matching step is unnecessary. Both Shi-Tomasi and LKT are implemented in the OpenCV library\added{~\cite{opencv}}.

\subsection{Simulation} 
\begin{figure}[t!]
  \centering
  \includegraphics[width=0.95\linewidth]{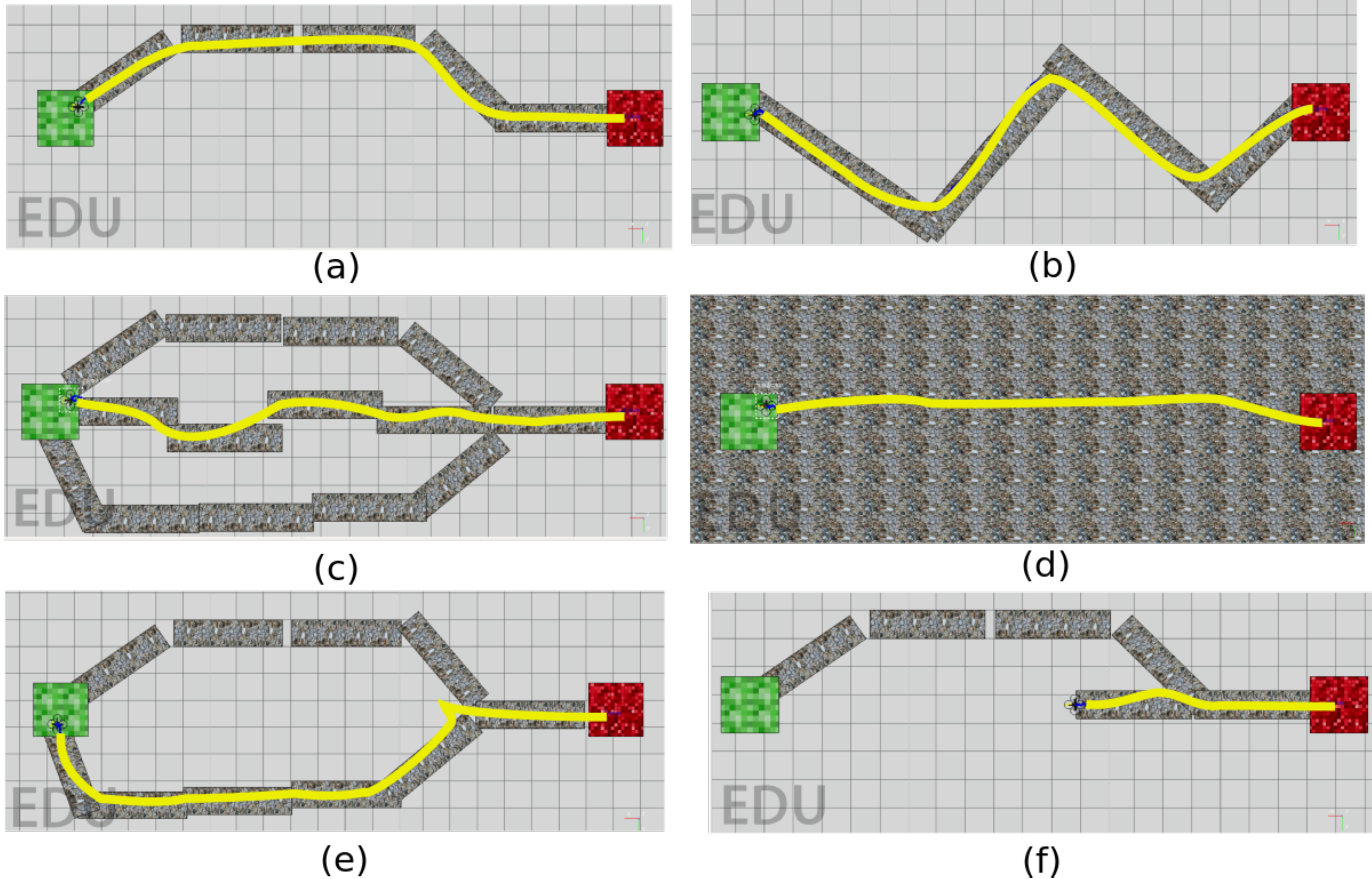}
  \caption{Simulations for evaluation of the proposed algorithm. The task consists in flying a 10 m length path from the red square (right) to the green square (left). (a-e) illustrate scenarios that have one or more feature-rich paths to the goal and (f) depicts a scenario where the proposed algorithm fails due to local minima. Parameters were kept constant: $\mathbf{p}_c=\mathbf{p}_o$, $\lambda = 0.4, \theta_{CS} = 60^{o.}, d=50$ pixel, $s=150$ pixel.}
  \label{fig:sil_results}
\end{figure}

The proposed solution was validated in different scenarios using the simulator V-REP by Coppelia Robotics. \deleted{Each scenario has at least one path to the goal which is rich in features. The values of $\lambda, \theta_{CS}, d, and s$ were similar throughout the simulations.} Fig.~\ref{fig:sil_results}(a-c) illustrate scenarios for which flying in a straight path to the goal is potentially risky for localization and map consistency. Therefore, in such scenarios, the proposed method is an asset, as it drives the vehicle through feature rich zones. Fig.~\ref{fig:sil_results}(d) considers a well textured scene. Although the potential-field framework does not benefit the navigation, it does not compromise reaching the goal. For 10 trials, the trajectory was $0.9 \pm 0.2$~m longer than flying in a straight line, which corresponds to  $9\%$ of the $10$~m straight path length. Fig.~\ref{fig:sil_results}(e-f) correspond to local-minima scenarios. The proposed method struggles to overcome a symmetric bifurcation (Fig.~\ref{fig:sil_results}(e)). \added{Such scenario with equally distributed features in opposite direction is not likely to happen in a real world environment}\deleted{, the vehicle finds a way out}. Finally, Fig.~\ref{fig:sil_results}(f) presents a scenario for which the method may eventually fail due to the local nature of potential field based solutions. Analyzing the simulations, if the goal velocity and the feature based velocity are pointing in opposite directions, the vehicle has reached a local minimum. This problem could be addressed by a high-level SLAM, which memorizes paths already travelled and selects an alternative path to the goal. If SLAM is not available -- due memory or computational power constraints -- increasing the value of $\lambda$ would enable to overcome the local minimum.

\begin{figure}[t!]
    \centering
  \includegraphics[width=0.95\linewidth]{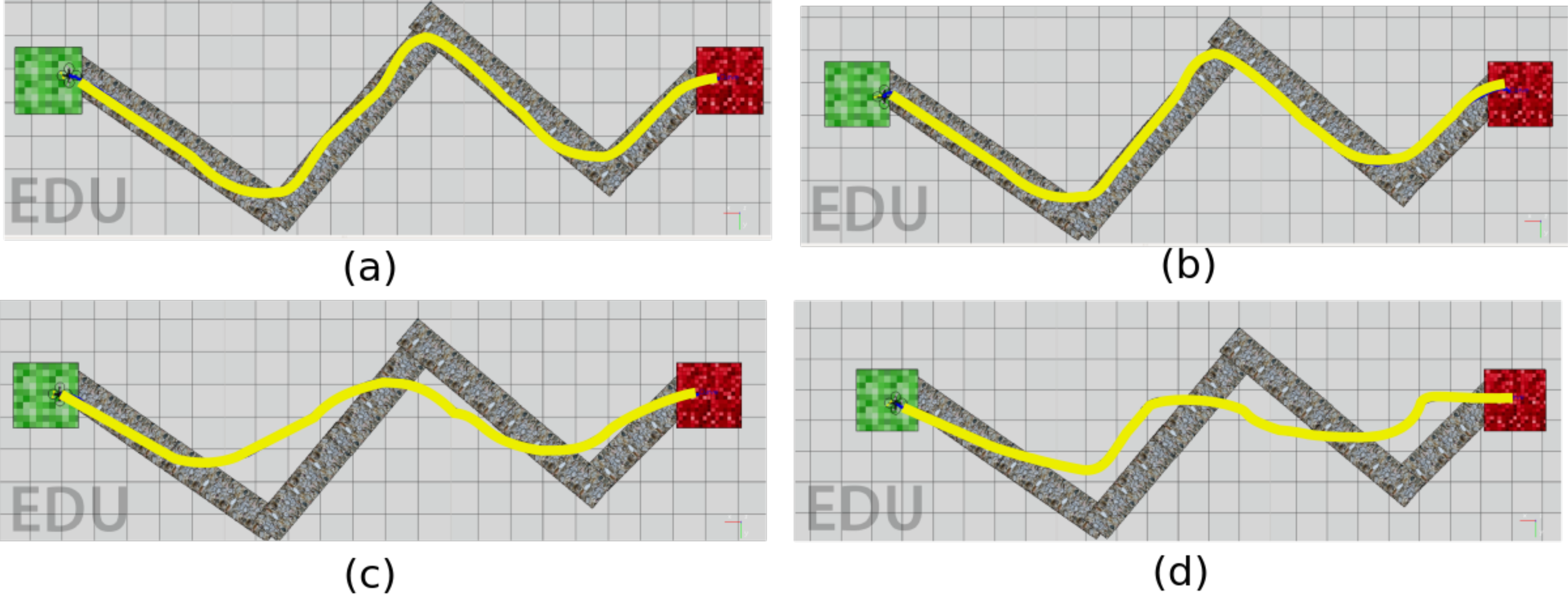}
  \caption{Effect of parameter $\lambda$: (a) $\lambda = 0.3$, (b) $\lambda = 0.5$, (c) $\lambda = 0.6$ and (d) auto-tuning\added{, on the experimental setup of Fig.~\ref{fig:sil_results}(b)}.}
  \label{fig:sil_lambda}
\end{figure}
\begin{figure}[t!]
\centering
  \includegraphics[width=0.85\linewidth]{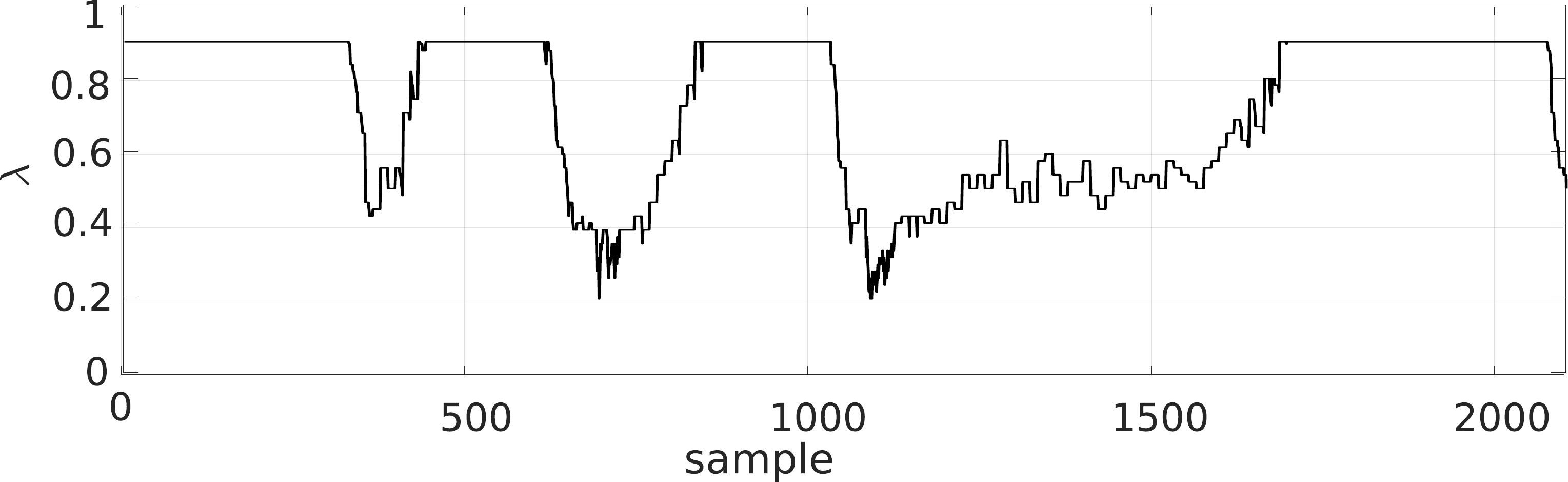}
  \caption{Value of parameter $\lambda$ using auto-tuning technique. Path travelled is as shown in Fig. \ref{fig:sil_lambda}(d).}
  \label{fig:lambda_autotuning}
\end{figure}

Figure~\ref{fig:sil_lambda} addresses the effect of varying $\lambda$. If $\lambda$ is low (Fig.~\ref{fig:sil_lambda}(a)) the vehicle takes a path more conservative with respect to features. In contrast, as $\lambda$ increases, the vehicle follows a trajectory closer to the straight line that links the start and goal positions (Fig.~\ref{fig:sil_lambda}(c)). An evaluation of the auto-tuning technique described in Section \ref{subsection:autotuning} is shown in Fig.~\ref{fig:sil_lambda}(d). $T_{f_{MAX}}$ and $T_{f_{MIN}}$ were set to $90\%$ and $25\%$ of the maximum number of features extracted per frame, respectively. The value of $\lambda$ throughout the simulation \added{of Fig.~\ref{fig:sil_lambda}(d)} can be seen in Fig.~\ref{fig:lambda_autotuning}. Online tuning had a positive effect on the overall performance of the task. In Fig.~\ref{fig:sil_lambda}(a) the vehicle travelled $12.05$ m to the goal\replaced{, while}{. Meanwhile,} using the auto-tuning technique, the path travelled was $11.1$ m length.

% \begin{figure}[t!]
% \centering
%   \includegraphics[width=0.65\linewidth]{distance_to_goal}
%   \caption{Real distance to goal upon estimated arrival. A total of 30 simulations were performed in a scenario as shown in Fig.~\ref{fig:sil_results}(a).}
%   \label{fig:distance2goal}
% \end{figure}

A total of 30 simulations using the scenario depicted in Fig.~\ref{fig:sil_results} were performed. For half of these simulations, the potential field framework was turned off. For the other half, it was turned on. In the first situation, vision-based localization failed after flying a fourth of the the straight line path. Afterwards, within an EKF framework, state estimation was purely based on inertial measurements propagation. IMU covariance error was extracted from the bagfile described in \cite{c3}. Meanwhile, the proposed method ensured visual localization worked properly during the entire flight. For both methodologies, based on its state estimation, the vehicle eventually reached the estimated goal position. For each trial, the distance between the final position vehicle and the real goal position was measured. \replaced{While when using the proposed active pipeline our method reached a distance of $0.85\pm 0.48$~m, the passive technique got $2.33\pm 1.02$~m,  showing that the proposed technique improves the overall location estimation.}{As shown in Fig. \ref{fig:distance2goal}, assuring visual pipelines does not fail boost state estimation performance.}\deleted{Notice that such comparison is only possible on simulations, as flying a real quadrotor based purely on inertial data is not a reliable approach.}

%The same set of parameters were employed for all evaluations.

\subsection{Experimental Setup with Real Data}

\begin{figure}[t!]
  \centering
  \includegraphics[width=0.9\linewidth]{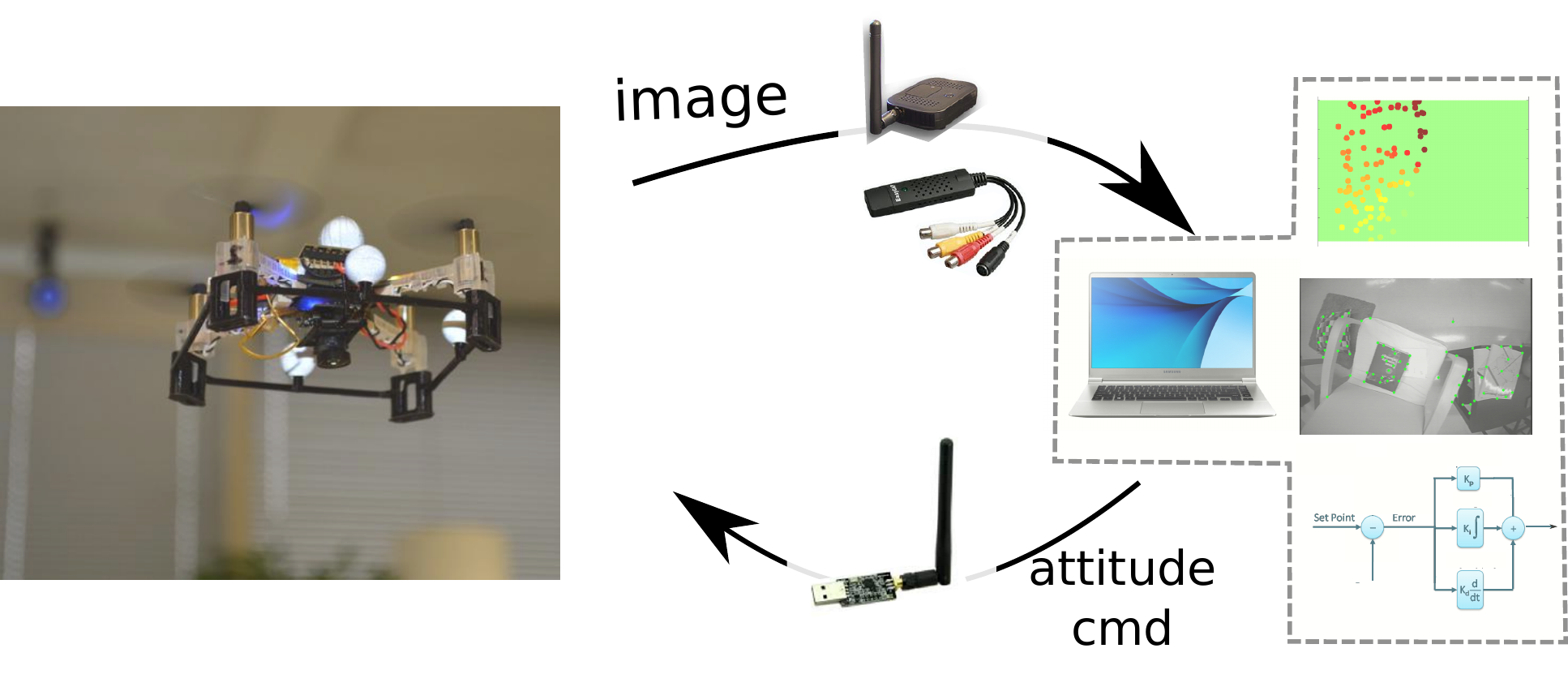}
  \caption{Experimental setup. Clockwise, starting from the top: the crazyflie equipped with the mini transmitter camera module FX798T, image receiver, and capture card; a notebook that runs the code; and an antenna for the communication with the vehicle.}
  \label{fig:experimental_setup}
\end{figure}

Figure~\ref{fig:experimental_setup} presents the main components employed in the experimental validation of the proposed method. The algorithm was tested in the mini-quadrotor Crazyflie 2.0\footnote{https://www.bitcraze.io/crazyflie-2/}, manufactured by BitCraze\replaced{,}{. This $27\,g$ open source vehicle allows a maximum take-off weight of $42$~g. It is} equipped with accelerometers, gyroscopes, magnetometer, pressure sensor, bluetooth, and a low-latency/long-range radio that allows for the robot to be operated (and log data) from a remote PC. The manufacturer's radio - Crazyradio - interfaces communication with a notebook, which computes the proposed action - collective thrust and attitude commands. \deleted{are sent to the vehicle using the package developed in}\cite{hoenig15}. \deleted{A customized PID controller ensures that the vehicle follows the desired reference velocity.} We attached to the vehicle a $4.7$~g mini transmitter camera module FX798T equipped with $120^o$ field of view lens - the camera faces downward. \deleted{This module broadcasts images using an embedded $5.8$~GHz transmitter.}  The frame-rate is $30$~Hz for a $720\times 480$ image resolution. The proposed algorithm runs off-board\deleted{ computer at frame-rate speed}. The images are transmitted via radio. Although variable, the measured delay upon image reception on the remote computer is about $150$~ms. The flying arena is equipped with an Optitrack\footnote{http://optitrack.com/} motion capture system\deleted{ manufactured by the company NaturalPoint. The system records the motion of retro-reflective markers using the optical-passive technique. Infrared cameras capture markers' position allowing}, \added{which estimates the} pose of the vehicle with millimetric precision. The motion capture system provides feedback for the control loop, while the feature-based potential field works online as the decision making process, sending desired velocity commands to the controller.
\begin{figure}[t!]
  \centering
  \includegraphics[width=0.85\linewidth]{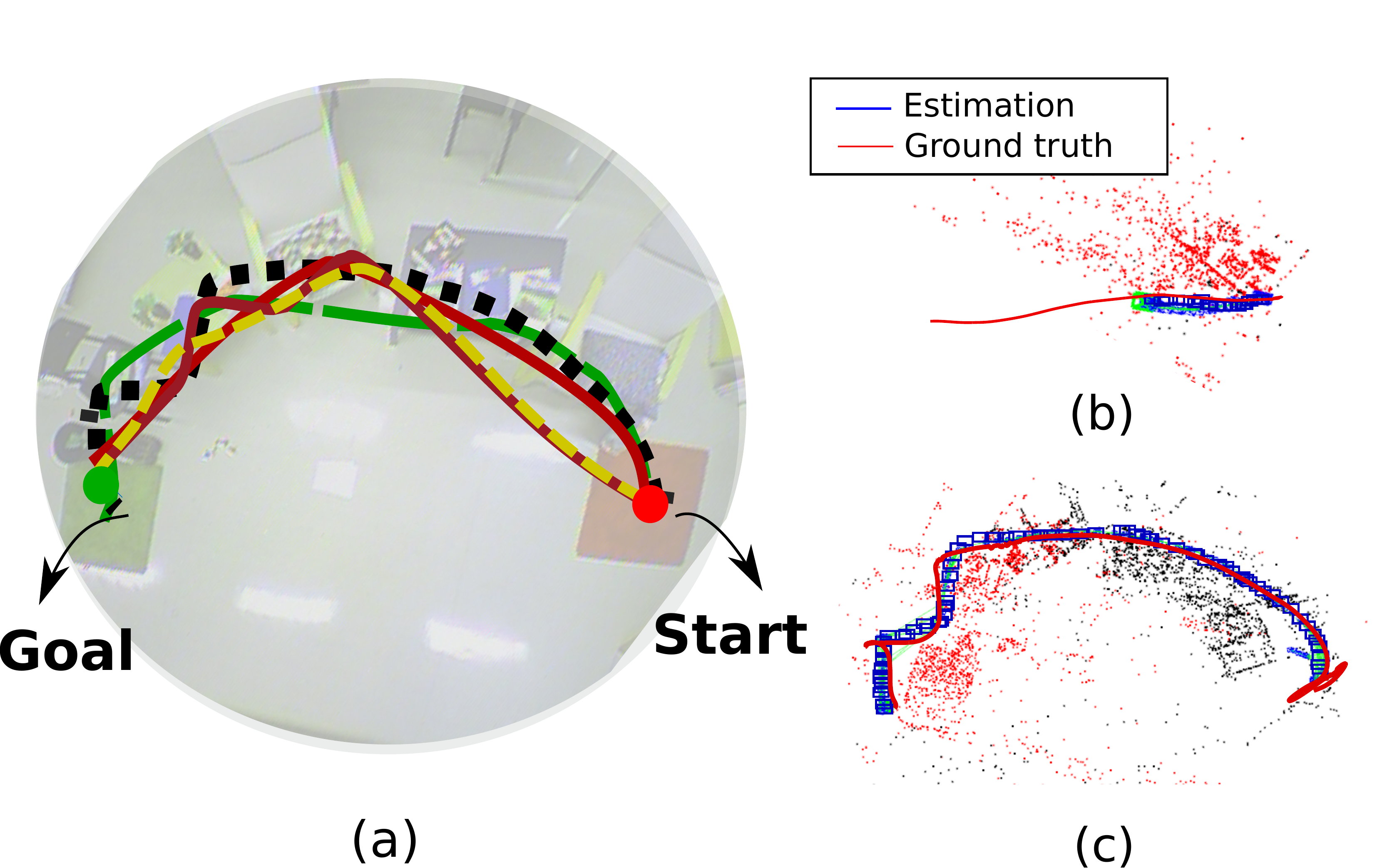}
  \caption{\replaced{Fig. (a) shows the first real scenario}{Scenario} for evaluation of proposed algorithm\deleted{(a)}. \replaced{Fig.~(b) shows that the localization}{Localization} fails in the absence of features \replaced{(passive technique)}{(b)}.\replaced{In Fig.~(c), we show results with our method, i.e. a feature}{Feature} based velocity vector drives the vehicle to the goal through a region rich in features\deleted{(c)}.}
  \label{fig:proof_of_concept}
\end{figure}
\subsection{Real Data Experiments Evaluation}
Multiple flights were performed in a $10\,\text{m}\times6\,\text{m}$ \added{home alike arena employed in European Robotics League (ERL)}\footnote{{https://www.eu-robotics.net/robotics\_league/erl-service/certified-test-beds/certified-test-beds.html}}\added{. In the first scenario,} shown in Fig.~\ref{fig:proof_of_concept}(a), the robot was autonomously controlled to fly from the starting position to the goal position as indicated by the two carpets. \deleted{As seen from the image, t}The straight path that links both carpets is a regular floor that does not contain good visual cues. The localization task for a trial is said to {\it succeed} if ORB-SLAM \cite{mur-artal15} manages to keep track of the pose of the vehicle and, therefore, a consistent map is built. A similar qualitative metric is employed in \cite{sadat14}. We stress again that one of the advantages of the proposed technique is that it can work under a SLAM framework, rather than on top (as most previous active perception strategies), ensuring a better localization and mapping result. Notice that we run ORB-SLAM offline, just for evaluation purposes, not for online localization. Multiple trials showed that taking the straight path towards the goal always results in localization failure due to the lack of visual features. Fig.~\ref{fig:proof_of_concept}(b) shows one of the trials where the localization estimation fails when a straight path is taken. Alternatively, using the proposed active method, the robot was always capable of reaching the goal while maintaining localization. Fig.~\ref{fig:proof_of_concept}(c) shows the path the robot takes, as well as the estimated state throughout the time in one of the experiments. Fig.~\ref{fig:proof_of_concept}(c) also shows the trajectory travelled for different trials with the proposed method where the robot reaches the goal while keeping track of its location.
%\begin{figure}[t!]
%  \includegraphics[width=0.90\linewidth]{multiple_trials}
%  \caption{\replaced{Different path for different trials of the problem defined in Fig.~\ref{fig:proof_of_concept}(a). We use the same evaluation parameters:} {Algorithm evaluation for} $\lambda=0.45$, $\hat{\theta}_{cs}=30^o$, and $\mathbf{p}_c = \mathbf{p}_o$.}
%  \label{fig:multiple_trials} 
%\end{figure}
\begin{figure}[b!]
  \centering
  \includegraphics[width=0.65\linewidth]{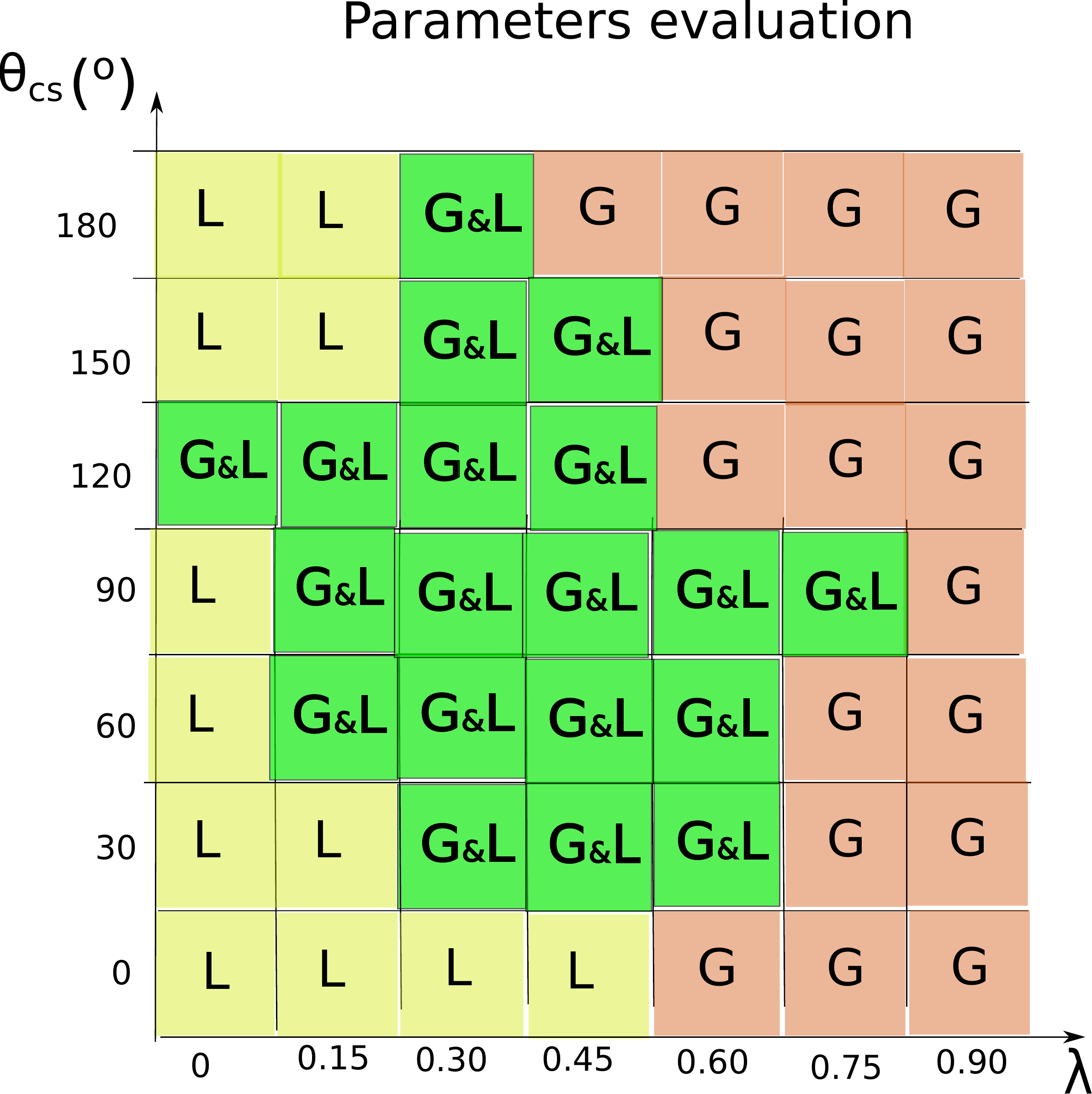}
  \caption{Algorithm evaluation for different values of $\lambda$ and $\hat{\theta}_{cs}$. \added{It was considered the experimental results presented in Fig.~\ref{fig:proof_of_concept}(a)}. $L$ means localization succeeded, $G$ means goal task succeeded, and $G\&L$ means both task succeed.}
  \label{fig:parameters}
\end{figure}

For the same scenario, the method was also evaluated for different values of $\lambda$ and $\hat{\theta}_{cs}$, while keeping $\mathbf{p}_c = \mathbf{p}_o$ constant. The goal task is said to $succeed$ if the vehicle reaches the goal within less than $0.15\,\text{m}$. Localization task success is defined as before. Results are shown in Fig.~\ref{fig:parameters}, where $L$ means localization task succeed, $G$ means the goal task succeed and $G\&L$ means both task succeed.\deleted{ There were a range of parameters values that led to success both tasks.} The localization task fails when the weight of the goal velocity vector is large, i.e. $\lambda$ is large, such that the vehicle takes a path similar to the straight one. Meanwhile, the goal task fails when the feature based potential field falls in a local minimum. Analyzing the image sequence, it is possible to observe that the vehicle hovers over a region that new features are not initialized and features contrary to the goal direction exert some attractive force. Consequently, the potential field force is not able to push the vehicle in the direction of the goal. As the value of $\hat{\theta}_{cs}$ increases, the attractive potential energy associated to features opposing the goal direction decreases. Therefore, the vehicle is no longer pulled backwards w.r.t. to the goal and reaches the final destination.

\begin{figure}[t!]
  \centering
  \includegraphics[width=0.7\linewidth]{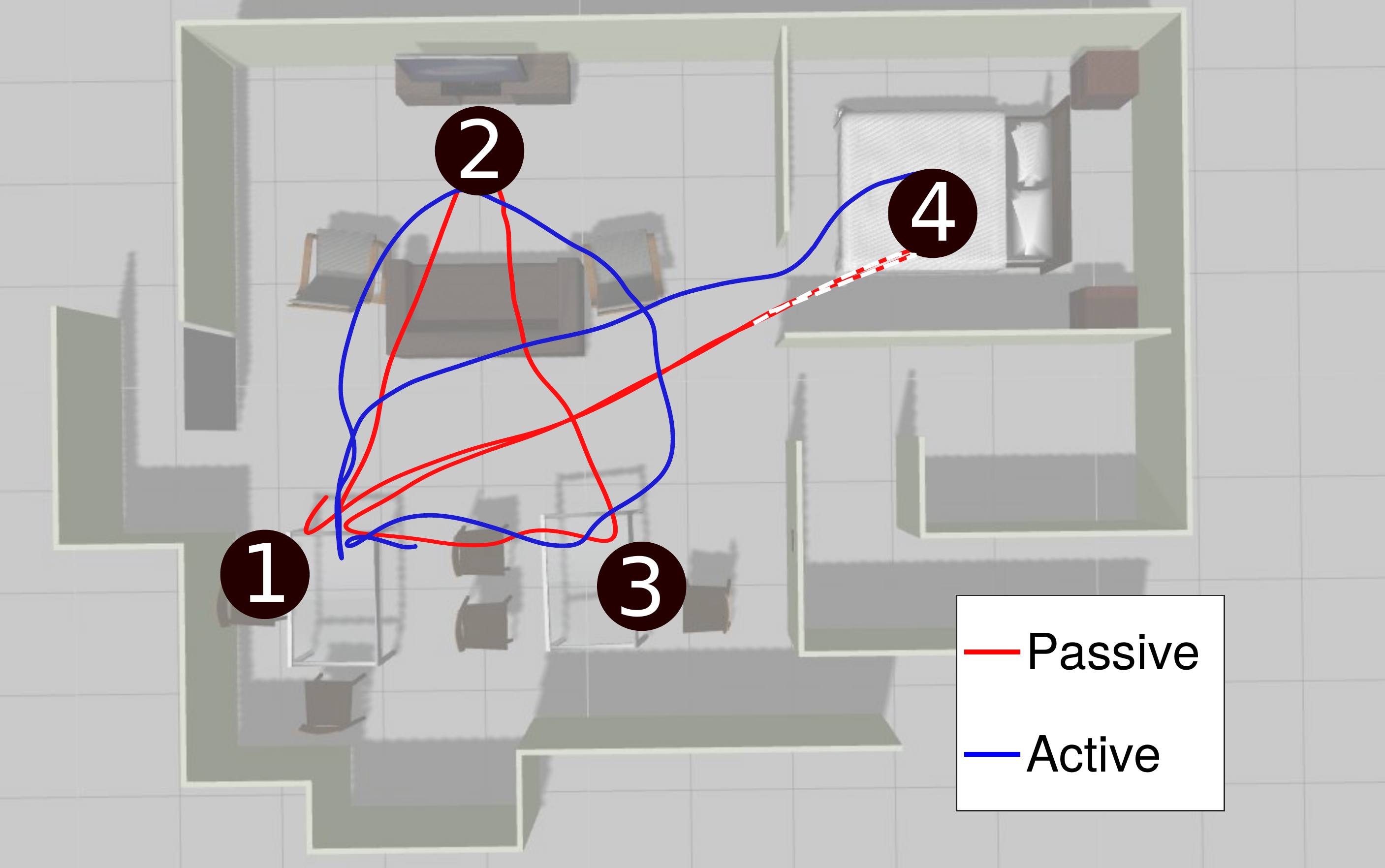}
  \caption{Passive and active flight on a home alike arena as employed in ERL. The vehicle visits waypoints in the order: 1, 2, 3, 1, and 4.}
  \label{fig:view_arena}
\end{figure}

\begin{figure}[t!]
  \centering
  \includegraphics[width=0.99\linewidth]{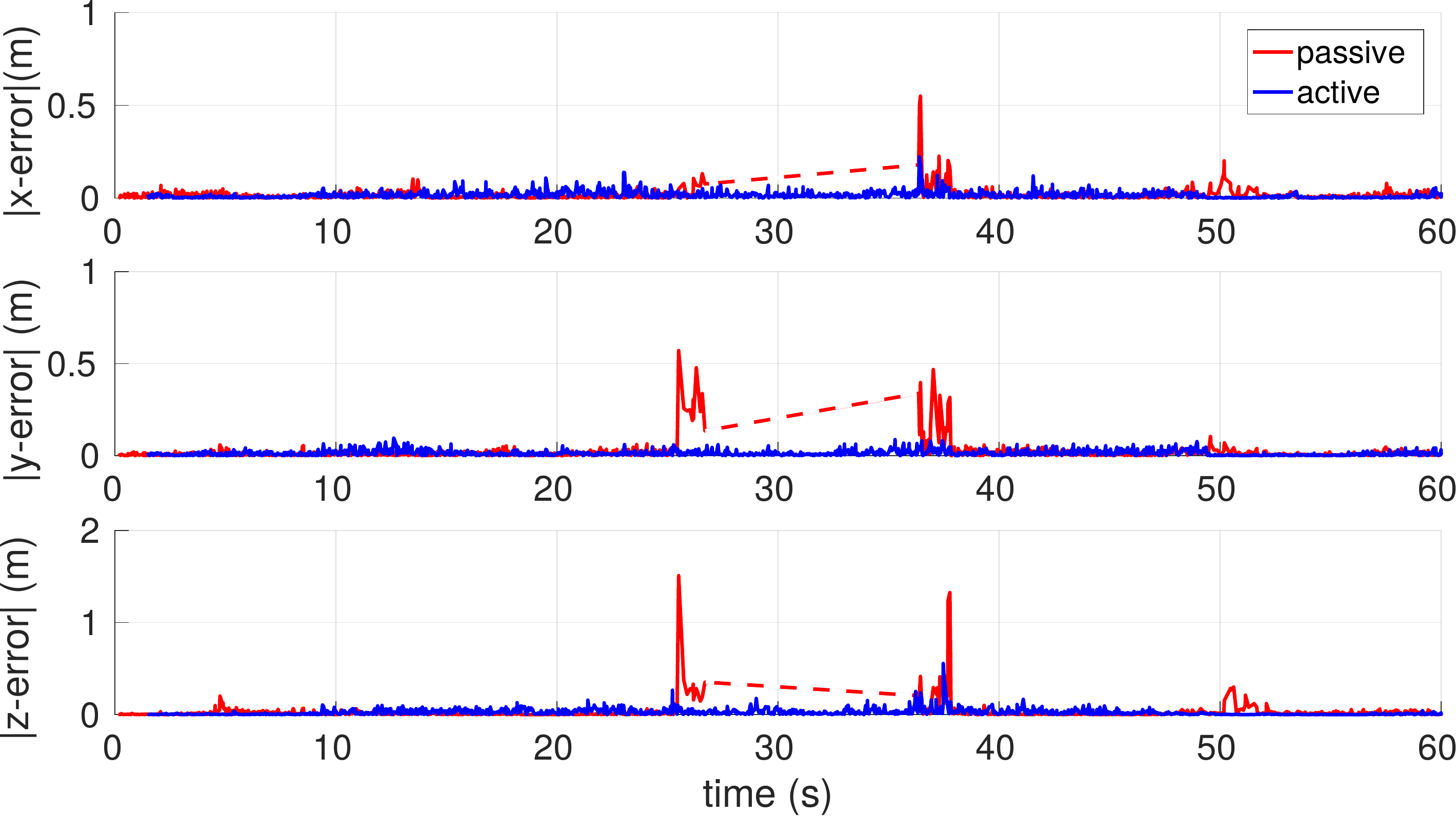}
  \caption{\added{Comparison of the visual odometry error (consecutive frames) for both the passive and active techniques using ORB-SLAM, for the experimental setup presented in Fig.~\ref{fig:view_arena}. The dash red line indicates that the robot is lost.}}
  \label{fig:quad_vo_error}
\end{figure}

\begin{figure}[t!]
  \centering
  \includegraphics[width=0.99\linewidth]{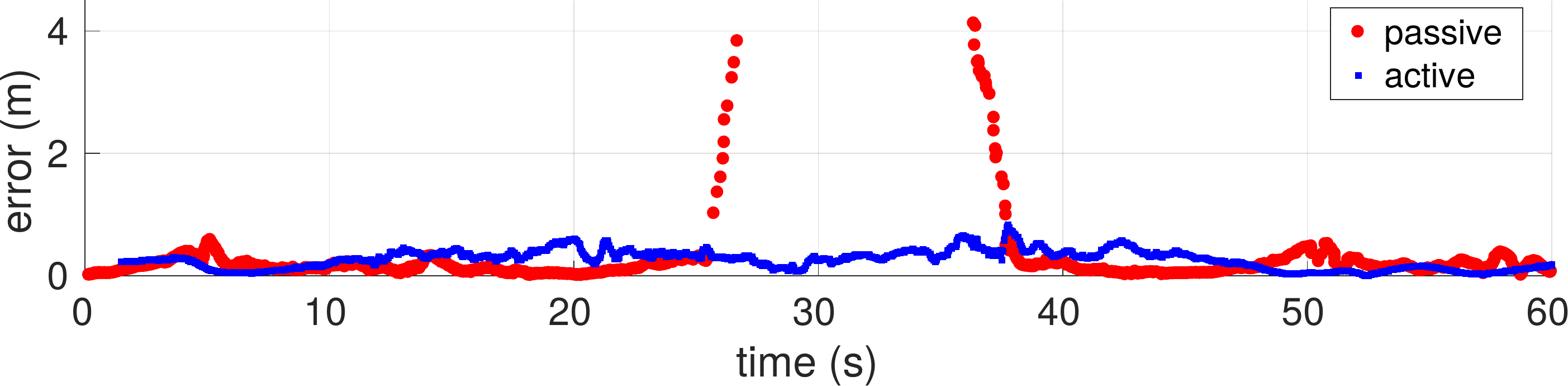}
  \caption{\added{Curves representing the SLAM estimated position error, for both passive and active techniques while doing the path presented in Fig.~\ref{fig:view_arena}. Points in the graph represent the interval of estimation.}}
  \label{fig:quad_slam_error}
\end{figure}

\begin{figure}[t!]
  \centering
  \includegraphics[width=0.97\linewidth]{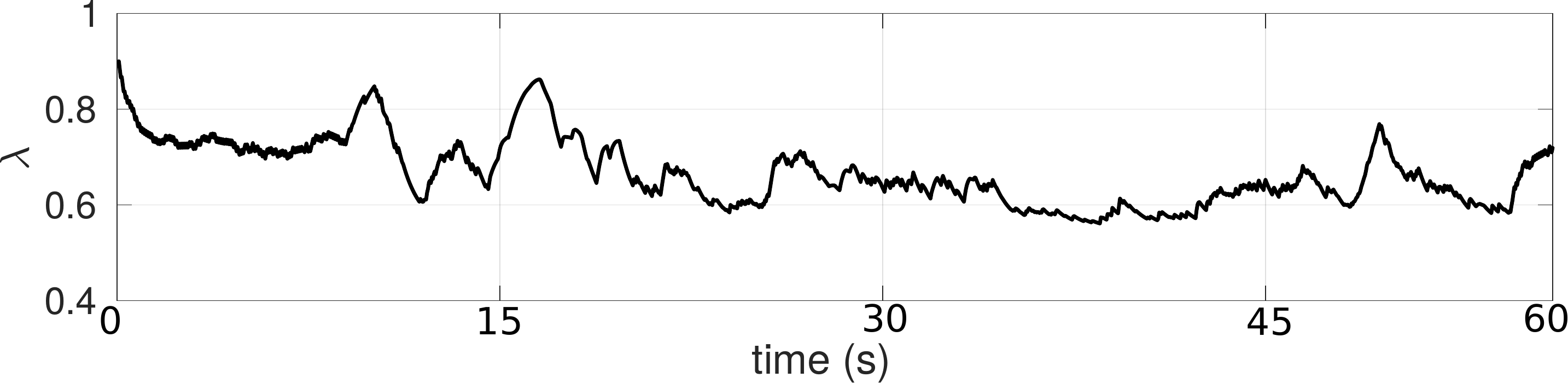}
  \caption{\added{Representation of the $\lambda$'s auto-tuning of the method presented in this paper, while making the path shown in Fig.~\ref{fig:view_arena} (blue curve).}}
  \label{fig:quad_lambda}
\end{figure}

\added{
To further evaluate the method in more realistic environments, we performed several long flights using all rooms of the home alike arena (Fig.~\ref{fig:view_arena}). The vehicle must visit waypoints (WP) selected by a human operator to explore different rooms of the apartment. As shown, the proposed method deviates the vehicle from straight paths that link consecutive waypoints. The VO performance is depicted in Fig.~\ref{fig:quad_vo_error} with (active mode) or without (passive mode) the proposed low level algorithm. The VO is obtained from ORB-SLAM, which requires proper 3D point initialization to propagate the relative scale. As long as enough visual cues are within the field of view of the camera, the error is small for both active and passive method. However, flying straight to the goal leads to poor feature regions closed to the bedroom entrance at $t=25~s$, where the passive estimator gradually fails to estimate the position. Before completely failing (dashed segment) visual odometry degrades quickly. The SLAM localization error (Fig.~\ref{fig:quad_slam_error}) presents a similar behaviour - error grows large before failing, but it is kept small when the map is concise. On the way out of the bedroom, the vehicle revisits some poorly initialized features leading to large errors during the recovering phase. 
If feedback was not provided by the motion capture system, such pose estimation failure would require to abort the mission and land in an open loop manner. 
In particular, for the active method flight tests, the value of $\lambda$ was computed using the auto-tuning method and it is shown in Fig.~\ref{fig:quad_lambda}. Overall, we performed about 150 meters of flight per method, corresponding to more than 8000 frames. With the proposed strategy, SLAM manages to estimate the pose for $99.1\%$ of the frames whereas only for $90.2\%$ was estimated using passive navigation. Therefore, thanks to the feature based strategy, VO and SLAM error are more likely to be kept small throughout the entire flight.}

% of the frames where estimated in the active method whereas only 90% was estimated for the passive method. 

% \begin{figure}[t!]
%   \centering
%   \includegraphics[width=0.99\linewidth]{vo_cum_error}
%   \caption{Visual odometry cumulative error (left) and, in particular, error with associated datapoints when vehicle loses track of features (right).}
%   \label{fig:vo_cum_error}
% \end{figure}

%When driven based on the goal velocity vector, i.e. in a straight path to the goal ($\lambda=1$), the localization pipeline fails (see ). Taking this setup into account, we run the proposed algorithm for the following parameters setting: $\mathbf{p}_c=\mathbf{p}_o$, $\lambda = 0.45$, $\theta_{cs} = 30^o$. Several trial for the same parameters setting is show Fig. \ref{fig:multiple_trials}. Although the vehicle takes different path, the vehicle reaches the goal and the localization task works correctly.

%In addition, an evaluation with the SLAM framework, for a set of different model parameters (from our method) are shown in the Tab.~\ref{}. From these results, one can see that the success of the ORB-SLAM (a state-of-the-art visual-SLAM technique) highly depend on the quality of the features {\color{blue}(describe the results, model parameters, in which the SLAM simply fails)}

\section{CONCLUSIONS}
% The control loop considers an additional reference velocity that points towards a feature rich region in the current image frame.
This paper proposes a simple, fast, low complex and map-independent solution for the active visual localization problem. The method, based on artificial potential fields, associates a potential energy to features that are classified as inliers in the current image frame. The goal direction is used to determine the corresponding charge intensity of these image features. Simulations and experimental results using a micro aerial vehicle, equipped with a downward looking camera, showed that the method can effectively drive the vehicle towards the goal, while avoiding no or poor featured regions. This active behaviour can greatly improve the localization performance and prevent common localization failures that are caused by low quality image features. The proposed active solution does not rely on a map and hence can be integrated within a SLAM framework to improve the accuracy and robustness of localization and mapping. Finally, the proposed method could benefit from the many techniques developed for potential fields, such random perturbations for escaping local minima.

%The proposed framework only considers attractive and neutral potential energy. As future work, the effect of associating repulsive energy to features classified as outliers shall be addressed. Also, reasoning about computing the feature based velocity vector in other point than the camera optical center is left as an open problem.

\end{document}